\documentclass[letterpaper]{article} % DO NOT CHANGE THIS
\usepackage[hyphens]{url}  % DO NOT CHANGE THIS
\usepackage{graphicx} % DO NOT CHANGE THIS
\usepackage{natbib}  % DO NOT CHANGE THIS AND DO NOT ADD ANY OPTIONS TO IT
\usepackage{caption} % DO NOT CHANGE THIS AND DO NOT ADD ANY OPTIONS TO IT
\usepackage{algorithm}
\usepackage{algorithmic}
\usepackage{amsmath}
\usepackage{amssymb}
\usepackage[preprint]{aaai2027}

\usepackage{newfloat}
\usepackage{listings}
\DeclareCaptionStyle{ruled}{labelfont=normalfont,labelsep=colon,strut=off} % DO NOT CHANGE THIS
\floatstyle{ruled}
\newfloat{listing}{tb}{lst}{}
\floatname{listing}{Listing}

\usepackage{booktabs}

\usepackage{multirow}
\usepackage[table]{xcolor}

\nocopyright 
\title{Structure-First Point Cloud Learning with Mapper Region Graphs} 

\author {
    Jeongbin You,
    Donggun Kim,
    Sejun Park,
    Seungsang Oh\textsuperscript{$\ast$}
}
\affiliations {
    Department of Mathematics, Korea University, Seoul 02841, South Korea\\
    \{dbwjdqls03, dorage1105, safespot, seungsang\}@korea.ac.kr\\
    \textsuperscript{$\ast$}Corresponding author
}

\begin{document}

\maketitle

\begin{abstract}
Robust point cloud classification has been studied through data augmentation, architectural redesign, and test-time adaptation. We investigate whether robustness can be improved by first organizing a point cloud into an overlapping region graph and subsequently propagating features over its region-intersection structure. Based on this principle, we propose {\bf Mapper-GNN}, a lightweight structure-first framework that constructs an input-dependent Mapper graph for each point cloud. Its nodes represent overlapping, locally coherent point subsets, while its edges encode region intersections. A shared pointwise MLP encodes region-relative and global coordinates into node features, which are subsequently processed by a residual multi-level GNN. We evaluate Mapper-GNN under a controlled protocol without corruption-specific augmentation, model ensembling, or test-time adaptation. With only 0.6M parameters, Mapper-GNN achieves the highest mean corruption accuracy among the evaluated models on ModelNet40-C, and remains competitive on ModelNet-C. We additionally evaluate Mapper-GNN on ScanObjectNN-C, extending the robustness analysis to real-world scanned objects. Its strongest results occur under density and noise corruptions, consistent with the motivation of reducing dependence on sampling-sensitive point neighborhoods. Mapper-GNN also performs favorably in extreme few-shot and reduced-point settings. These results demonstrate that Mapper-induced region graphs provide an effective structural inductive bias for robust and parameter-efficient point cloud classification.
\end{abstract}

\section{Introduction}

Point cloud classification has advanced through increasingly expressive architectures that aggregate geometric information from individual points or sampled representatives~\cite{pointnet2017,pointnetpp,dynamicgcnn2019,pct}.
Despite strong performance on clean benchmarks, these models can be sensitive to changes in the input sampling pattern.
Variations in point density, positional noise, and missing points may alter the local neighborhoods and pairwise relations used to construct point features, even when the underlying object remains recognizable.
Consequently, improved clean accuracy does not necessarily translate into greater corruption robustness, as demonstrated by ModelNet40-C~\cite{modelnet40c} and ModelNet-C~\cite{pointcloudc}.

\begin{figure}[t]
\centering
\includegraphics[width=\columnwidth]{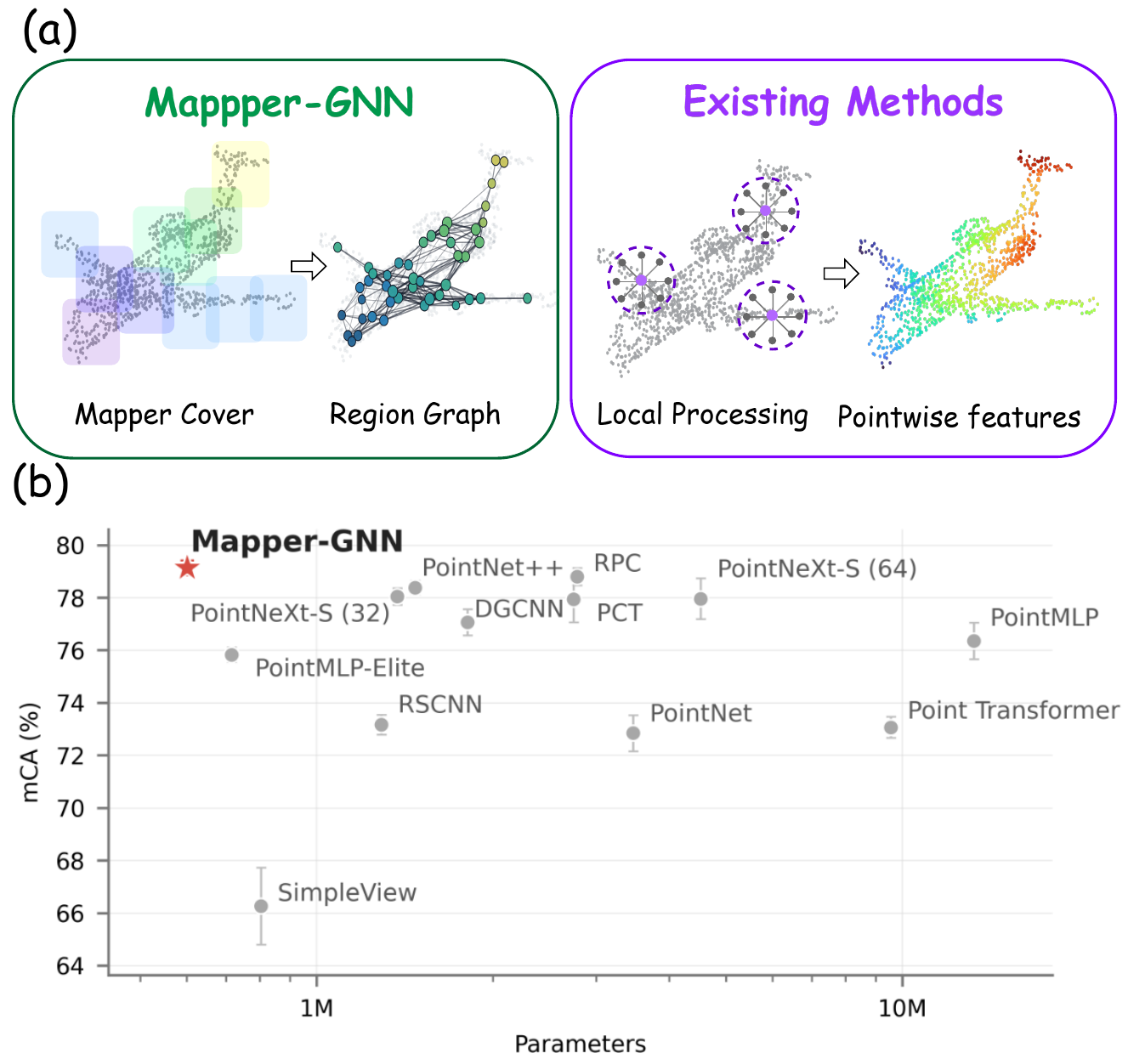}
\caption{Structure-first point cloud learning with Mapper-GNN.
(a) Mapper-GNN first constructs an overlapping region graph, while existing methods directly learn point-level features.
(b) Mean corruption accuracy (mCA) versus parameter count on ModelNet40-C.}
\label{fig1}
\end{figure}

Existing robustness approaches primarily modify the training distribution, redesign point-based architectures, remove unreliable points, or adapt models at test time. We investigate a complementary question: \emph{can robustness be improved by first constructing a region graph-based structural representation before neural feature propagation?}

We refer to this design principle as \textbf{structure-first point cloud learning}. Instead of propagating features directly over points or sampled neighborhoods, we first construct an input-dependent overlapping region graph using Mapper~\cite{mapper2007}. Each node represents a locally coherent point subset, and two nodes are connected when their associated subsets intersect. The resulting region-intersection graph serves as the domain for subsequent feature propagation.

This representation is motivated by the sampling sensitivity of point-level neighborhoods. Small perturbations, point removal, or density variation can substantially change nearest-neighbor relations. In contrast, region-level summaries and their overlap structure may remain more consistent when the coarse organization of the object is preserved. We therefore hypothesize that a Mapper-induced region graph can provide a useful structural inductive bias, particularly under density and noise corruptions.

Based on this idea, we propose {\bf Mapper-GNN}, a lightweight point cloud classification framework built around a Mapper-induced region graph. Within each region, a shared pointwise MLP encodes both region-relative and global coordinates, and max pooling produces an initial node representation. A residual GNN then propagates features over the region-intersection graph. Multi-level node features are aggregated using Jumping Knowledge-style fusion~\cite{pmlr-v80-xu18c} and combined with a direct pooled representation of the initial region features for classification.

We evaluate Mapper-GNN under a controlled protocol without corruption-specific augmentation, model ensembling, or test-time adaptation, thereby isolating the contribution of the region-level structural abstraction. Experiments across ModelNet40-C, ModelNet-C, and the
real-world ScanObjectNN-C benchmark demonstrate competitive
corruption robustness with substantially fewer parameters. The largest improvements are observed under density and noise corruptions, consistent with our motivation of reducing dependence on unstable point-level neighborhoods. Additional experiments under extreme few-shot training and reduced input-point settings further demonstrate the parameter and data efficiency of the proposed representation.

Our contributions are summarized as follows:
\begin{itemize}
\item We introduce structure-first point cloud learning, in which an input-dependent Mapper graph provides an intermediate computational representation that shifts feature propagation from individual points to overlapping, locally coherent regions.
\item We develop Mapper-GNN, a lightweight architecture combining region-relative point-to-node encoding, residual multi-level graph propagation, Jumping Knowledge-style feature aggregation, and learnable base-to-graph residual fusion.
\item Through controlled evaluations on corruption benchmarks, we demonstrate that the proposed region-level abstraction provides strong parameter-efficient robustness under common corruptions, extreme few-shot learning, and global point sparsity, with particularly favorable performance under density and noise perturbations.
\end{itemize}

\section{Related Work}

\paragraph{Point Cloud Classification.}
PointNet~\cite{pointnet2017} established direct learning on unordered point sets using shared pointwise transformations and permutation-invariant global pooling. 
PointNet++~\cite{pointnetpp} subsequently introduced a hierarchical local aggregation over progressively larger neighborhoods. 

Graph-based methods such as DGCNN~\cite{dynamicgcnn2019} and RSCNN~\cite{liu2019rscnn} explicitly model local geometric relations.
More recent architectures use attention-based interactions, as in PCT~\cite{pct} and Point Transformer~\cite{pointtransformer}, or strengthen hierarchical point processing through improved training and scaling strategies, as in PointNeXt~\cite{pointnext}.

Despite their architectural differences, such methods primarily operate on individual points or sampled representatives. 
In contrast, Mapper-GNN performs feature propagation over an overlapping region graph whose nodes represent locally coherent point subsets and whose edges encode explicit intersections between regions. Our contribution therefore lies in introducing a region-level structural representation rather than another point-level aggregation rule.

\paragraph{Robust Point Cloud Classification.}
Although modern point cloud classifiers perform well on clean benchmarks, their accuracy can degrade substantially under realistic corruptions. ModelNet40-C~\cite{modelnet40c} evaluates robustness under density, noise, and transformation corruptions, while ModelNet-C~\cite{pointcloudc} covers object-level and sensor-level corruptions. ScanObjectNN-C~\cite{Wang_2023_ICCV} complements these
CAD-derived benchmarks by evaluating common corruptions on
real-world scanned objects.
These benchmarks demonstrate that improved clean accuracy does not necessarily imply greater corruption robustness.

Existing approaches improve robustness through several complementary strategies. 
PointMixup~\cite{pointmixup}, RSMix~\cite{rsmix}, and PointCutMix~\cite{pointcutmix} construct augmented samples by mixing points or rigid subsets from different objects. 
ModelNet40-C further investigates combinations of augmentation and test-time adaptation, whereas the ModelNet-C study introduces WOLFMix and the Robust Point Cloud Classifier (RPC). 
PointCVaR~\cite{li2024pointcvar} removes harmful outliers, while PointDGMamba~\cite{yang2025pointdgmamba} and PointDGRWKV~\cite{yang2026pointdgrwkv} improve generalization to unseen domains. Topology-preserving augmentation has also been explored to generate diverse samples while maintaining their underlying manifold structure~\cite{bi2026shaping}.

In contrast, we isolate the effect of structural abstraction under a controlled protocol without corruption-specific augmentation, model ensembling, or test-time adaptation. We examine whether replacing sampling-sensitive point neighborhoods with a Mapper-induced region graph can provide an effective structural prior for robustness.

\begin{figure*}[t] \centering \includegraphics[width=\linewidth]{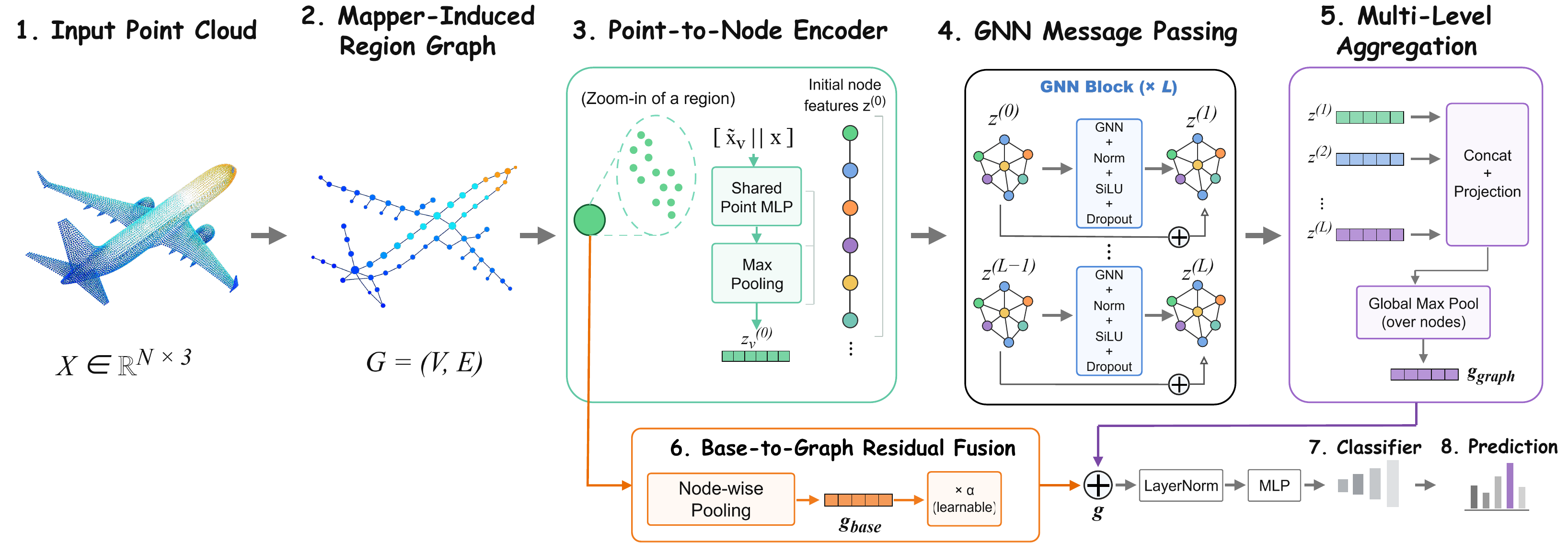} \caption{Mapper-GNN architecture} \label{fig3} \end{figure*}

\paragraph{Topological Data Analysis and Mapper.}
Topological Data Analysis (TDA) has been integrated with neural networks mainly through persistent-homology-based representations.  
Early work introduced learnable layers for persistence diagrams~\cite{hofer2017topological}, while PersLay~\cite{carriere2018perslay} provided a learnable and permutation-invariant framework for aggregating persistence features. 
More recent studies have explored adaptive filtration learning for point
clouds~\cite{adaptiveph} and point-level topological representations that
complement conventional geometric features~\cite{topf}.

Mapper~\cite{mapper2007} provides a graph-based alternative to persistence-diagram representations.
Given a lens function and an overlapping cover, it clusters the corresponding pullback sets and connects clusters that share samples. 
The resulting graph summarizes how locally coherent regions overlap and connect.
Mapper is closely related to the Reeb graph, which captures how connected components of lens level sets split and merge over a continuous domain. 
This connection has motivated theoretical studies of Mapper approximation and statistical consistency~\cite{carriere2018structure,carriere2018statistical}, including extensions to multivariate and data-dependent filters such as PCA-based lenses~\cite{carriere2022multivariate}.

Recent Mapper variants adapt the construction itself.
G-Mapper~\cite{gmapper} learns a data-dependent cover, while Differentiable
Mapper~\cite{differentiablemapper} enables optimization of the lens function. 
Our work instead constructs a non-learned, input-dependent Mapper graph as an intermediate representation for each point cloud. 
Unlike persistence-based approaches that provide topological summaries as auxiliary features, Mapper-GNN performs feature propagation directly over the region-intersection graph.
Classical Mapper consistency results do not directly apply to our multivariate lens and DBSCAN-based clustering procedure. We therefore regard them as motivation for using Mapper as a coarse structural representation that can remain insensitive to moderate sampling perturbations. 
Our experiments evaluate whether this region-level abstraction remains sufficiently stable under common point cloud corruptions to improve classification robustness.

\section{Method} 

We investigate whether a topology-inspired structural representation can improve the robustness of point cloud classification without relying on a heavy backbone or specialized geometric modules. To this end, we propose {\bf Mapper-GNN}, which combines a Mapper-induced region graph with lightweight point-to-node encoding and graph neural network-based message passing. The complete computation pipeline is illustrated in Figure~\ref{fig3}.

\subsection{Mapper-Induced Region Graph}

Let $X=\{x_i\}_{i=1}^N \subset \mathbb{R}^3$ denote an input point cloud. Following the Mapper construction~\cite{mapper2007}, we build a region graph $G=(V,E)$ by sequentially applying a lens function, an overlapping cover of the lens space, and clustering within each pullback set. We use Mapper to construct a region-level computational scaffold for robust feature propagation.

\begin{figure}[ht] \centering \includegraphics[width=\columnwidth]{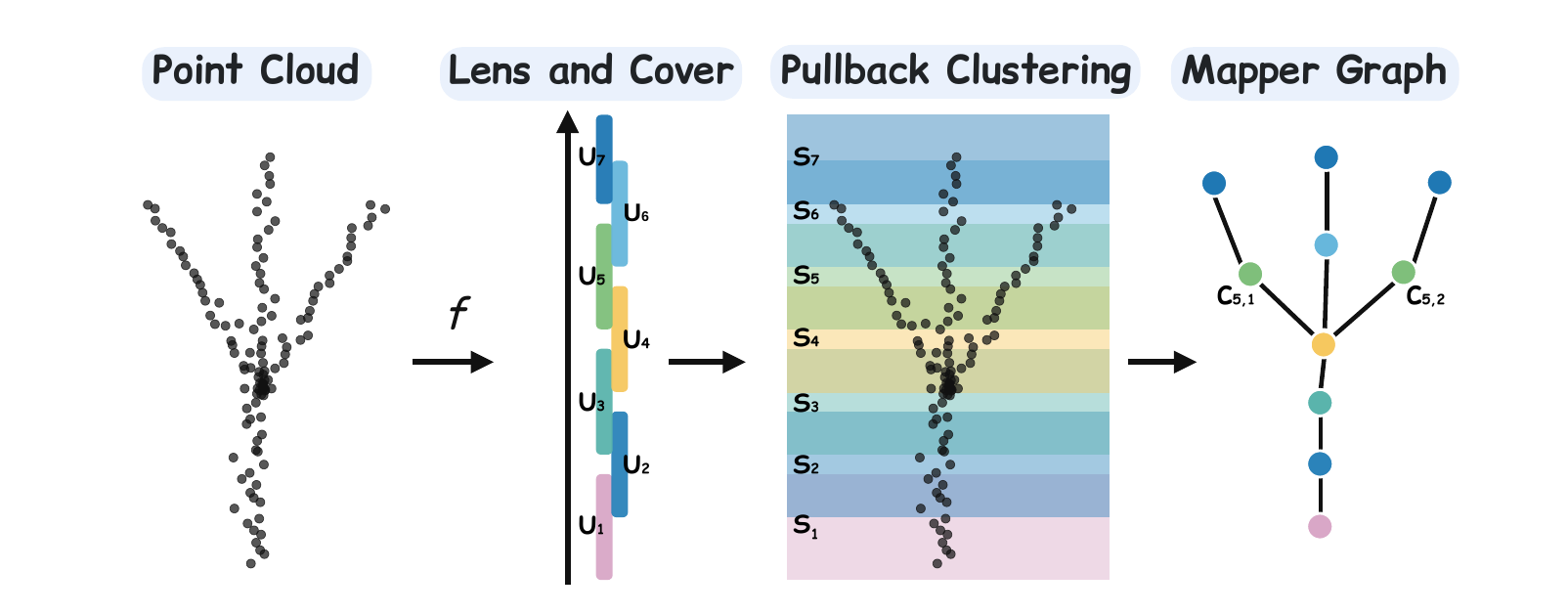} \caption{Mapper with a 1D lens; our model uses 3D identity/PCA lenses and a cubical cover.} \label{fig2} \end{figure} 

We consider two fixed lens functions, namely the identity lens and the PCA-aligned lens:
$$
f_{\mathrm{Id}}(x)=x,
\qquad
f_{\mathrm{PCA}_3}(x)=W^{\top}(x-\mu),
$$
where $\mu$ is the point-cloud centroid and the columns of $W$ are its three principal axes. The identity lens retains the original coordinates, whereas the PCA$_3$ lens provides a centered, principal-axis-aligned representation. Both lenses provide simple and computationally efficient coordinate systems for constructing regions. We then define an overlapping cubical cover $\mathcal{U} = \{U_i\}_{i\in\mathcal{I}}$ over the lens image $f(X) \subset \mathbb{R}^3$. The cover resolution controls region granularity, while the overlap allows points to belong to multiple regions.

For each cover element $U_i$, we extract the corresponding
pullback $S_i=f^{-1}(U_i)\subset X$. Since a pullback may
contain multiple spatially separated components, we partition
$S_i$ using DBSCAN~\cite{ester1996density} into clusters
$\{C_{i,j}\}$.

Each non-noise cluster $C_{i,j}$ defines a graph node $v_{i,j} \in V$, whose associated point subset is $X_{v_{i,j}} = C_{i,j} \subset X$. Two nodes are connected whenever their associated clusters share at least one point:
$$(v_{i,j},v_{i'\!,j'}) \in E \quad \Longleftrightarrow \quad C_{i,j}\cap C_{i'\!,j'}\neq\emptyset.$$
Consequently, the graph edges explicitly encode intersections between locally coherent regions, producing an overlapping region-level representation of the input point cloud.

\subsection{Point-to-Node Encoder}

For each node \(v\in V\), we encode its associated sub-cloud $X_v \subset X$ into a node embedding. For each region, we compute its center and radius:
$$c_v = \frac{1}{|X_v|} \sum_{x\in X_v}x, \quad r_v = \max \Big( \max_{x\in X_v}\|x-c_v\|_2, \epsilon \Big),$$
where \(\epsilon>0\) prevents numerical instability. The point descriptor $h_{v,x}$ of each point $x$ in a sub-cloud $X_v$ is represented by using both a normalized local coordinate (region-relative) and a global coordinate of $x$:
$$h_{v,x} = \phi \left( \left[\, \widetilde{x}_{v} \Vert\, x\, \right] \right) \text{ \ with\, } \widetilde{x}_{v} = \frac{x-c_v}{r_v},$$
where $\phi\!:\! \mathbb{R}^{6}\rightarrow\mathbb{R}^{D}$ is a shared lightweight pointwise MLP and \(\Vert\) denotes feature concatenation. 
This parameter-free normalization provides a translation- and scale-normalized description of the geometry within each region without introducing additional complexity.

The initial node embedding is obtained by node-wise max pooling:
$$z_v^{(0)} = \max_{x\in X_v} h_{v,x}.$$
The local coordinates encode the internal geometry of each region, while the global coordinates preserve its position within the original point cloud.

\begin{figure}[t]
\centering
\includegraphics[width=\columnwidth]{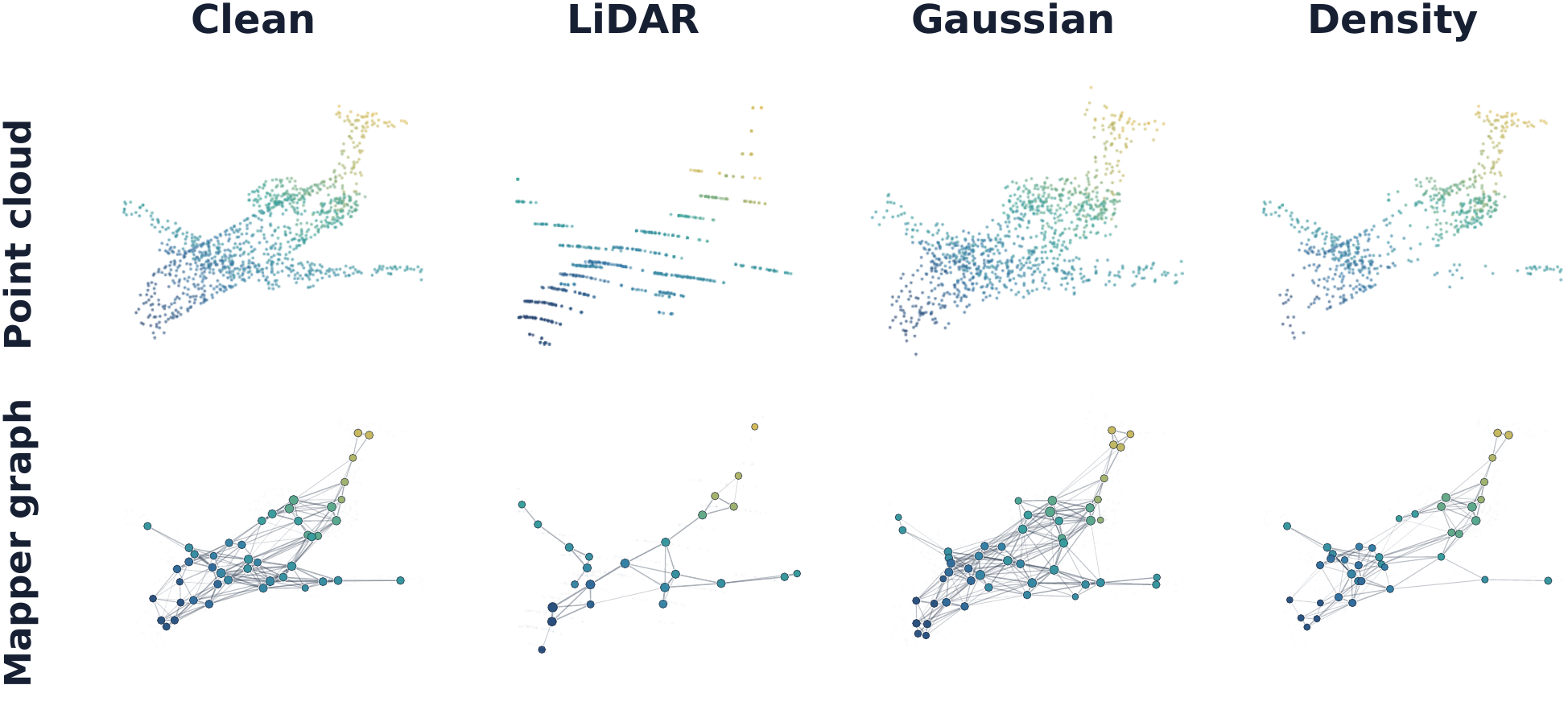} 
\caption{Corrupted point clouds and their Mapper graphs.}
\label{fig4}
\end{figure}

\subsection{GNN Message Passing}
Starting from the initial node features 
\(Z^{(0)}=\{z_v^{(0)}\}_{v\in V}\), we apply \(L\) message passing blocks over the Mapper-induced graph. At layer \(\ell\), the graph operator computes
$$\hat{z}_v^{(\ell)} = \operatorname{GNN}^{(\ell)} \Big( z_v^{(\ell-1)}, \big\{ z_u^{(\ell-1)} : u\in\mathcal{N}(v) \big\} \Big),$$
where $\mathcal{N}(v)$ denotes the neighbors of $v$. We instantiate $\operatorname{GNN}^{(\ell)}$ using GCN~\cite{gcn2017}, GAT~\cite{gat2018}, or GIN~\cite{gin2019}. Each block applies graph normalization~\cite{cai2021graphnorm}, SiLU activation~\cite{elfwing2018sigmoid}, dropout~\cite{srivastava2014dropout}, and residual propagation:
$$z_v^{(\ell)} = z_v^{(\ell-1)} + \operatorname{Dropout} \big( \operatorname{SiLU} \big( \operatorname{GraphNorm} \big( \widehat{z}_v^{(\ell)} \big) \big) \big).$$
During training, we apply DropEdge~\cite{rong2020dropedge} with a rate of $0.3$.

To preserve representations produced at different message-passing
depths, we use Jumping Knowledge-style layer
aggregation~\cite{pmlr-v80-xu18c}. The graph representation is obtained by concatenating and projecting the intermediate node features from all layers, followed by global max pooling:
$$g_{\mathrm{graph}} = \max_{v\in V} \Big( \psi_{\mathrm{multi}} \big( z_v^{(1)} \Vert z_v^{(2)} \Vert \cdots \Vert z_v^{(L)} \big) \Big),$$
where $\psi_{\mathrm{multi}}\!:\! \mathbb{R}^{LD}\rightarrow\mathbb{R}^{D}$ is a learnable projection.

\subsection{Base-to-Graph Residual Fusion}

In parallel with graph propagation, we aggregate the initial node
features before message passing. The resulting base representation is
combined with the graph representation through learnable residual fusion:
\[
g = g_{\mathrm{graph}} + \alpha\, g_{\mathrm{base}},
\qquad
g_{\mathrm{base}} = \max_{v\in V} z_v^{(0)},
\]
where \(\alpha\) is a learnable scalar. The base representation provides a
direct path from the region-wise geometric encoder to the classifier. The
fused representation is normalized ~\cite{ba2016layer} and passed to a lightweight
classification head:
\[
p =
\operatorname{softmax}
\left(
\mathcal{H}_{\mathrm{cls}}
\left(
\operatorname{LayerNorm}(g)
\right)
\right),
\]
where \(\mathcal{H}_{\mathrm{cls}}\) consists of an MLP followed by a linear
output layer. Let \(p_y\) denote the predicted probability assigned to the
ground-truth class \(y\). The network is optimized using the standard
cross-entropy objective:
\[
\mathcal{L}_{\mathrm{cls}} = -\log p_y.
\]

\section{Experiments}
\subsection{Experimental Setup}
To evaluate Mapper-GNN, we consider three complementary settings: 
(1) robustness to common point cloud corruptions, 
(2) few-shot classification on clean and corrupted data, and 
(3) robustness to global point sparsity.

\paragraph{Datasets.}
ModelNet40~\cite{wu20153d} is a widely used benchmark for 3D object classification, consisting of CAD-derived point clouds from 40 object categories. ModelNet40-C~\cite{modelnet40c} extends ModelNet40 with 15 realistic corruption types at five severity levels. It evaluates corruption robustness by training models on clean ModelNet40 and testing them on corrupted point clouds.

We use ModelNet40-C as our primary corruption benchmark and additionally evaluate on ModelNet-C~\cite{pointcloudc} and ScanObjectNN-C~\cite{Wang_2023_ICCV}.
While the former provides an independent CAD-based corruption protocol, ScanObjectNN-C evaluates robustness on real-world scanned point clouds.

\paragraph{Evaluation Protocols.}
As highlighted by SimpleView~\cite{goyal2021revisiting}, 3D point cloud classification results can depend on
protocol-level choices such as data augmentation, model selection, test-time voting, and point resampling. We therefore keep the main training and evaluation settings consistent across models.
For ModelNet40-C and ModelNet-C, all models are trained on clean ModelNet40 using $N=1024$ input points. For ScanObjectNN-C, models are trained on the corresponding clean ScanObjectNN split. Across all benchmarks, models use point coordinates only, without surface normals or additional geometric attributes. We do not use test-time voting, model
ensembling, aggressive data augmentation, or label smoothing.
Unless otherwise specified, models are trained with a batch size of 128
using Adam~\cite{kingma2014adam}, an initial learning rate of
\(10^{-3}\), and a weight decay of \(10^{-4}\). The learning rate is
decayed by a factor of 0.9 every 10 epochs using StepLR. Following standard corruption-evaluation practice~\cite{modelnet40c},
we select checkpoints based on the corresponding clean-set accuracy and evaluate them unchanged under all corruption types and severity
levels. For the point-sparsity experiments, we reuse the checkpoint selected on clean ModelNet40.

\paragraph{Evaluation Metrics.}
For ModelNet40-C, \textbf{m}ean \textbf{C}orruption \textbf{A}ccuracy (\textbf{mCA}) is computed by averaging accuracy over five severity levels for each corruption and then across all 15 corruption types. DEN, NOI, and TRA denote the corresponding averages within the density, noise, and transformation groups, respectively.
\textbf{Retention} measures the percentage of clean accuracy preserved under
corruption. For ScanObjectNN-C, it is computed as the mean accuracy
across corruption types divided by the corresponding clean accuracy.
For the few-shot experiments, we compute $\mathrm{mCA}/\mathrm{cOA}$
for each $n$-way $m$-shot setting and report the average over all six settings.

\begin{table}[h]
\centering
\small
\setlength{\tabcolsep}{2.0pt}
\renewcommand{\arraystretch}{1.0}
\begin{tabular}{lc|c|ccc}
\toprule
\textbf{Model} & \textbf{Params} $\downarrow$ & \textbf{mCA} $\uparrow$ & DEN & NOI & TRA \\
\midrule
PointNet & 3.47M & 72.84 & \ 68.52 \ & 73.37 & \ 76.64 \ \\
PointNet++ & 1.47M & 78.38 & 69.04 & 84.32 & 81.80 \\
DGCNN & 1.81M & 77.07 & 68.92 & 76.34 & 85.95 \\
RSCNN & 1.29M & 73.17 & 64.44 & 77.37 & 77.71 \\
PCT & 2.74M & 77.93 & 69.27 & 79.17 & 85.36 \\
SimpleView & 0.80M & 66.27 & 52.78 & 74.33 & 71.68 \\
Point Transformer & 9.55M & 72.04 & 61.99 & 77.84 & 76.28 \\
PointNeXt-S ($C=32$)\! & 1.37M & 78.04 & 69.19 & 81.69 & 83.24 \\
PointNeXt-S ($C=64$)\! & 4.52M & 77.96 & 69.15 & 81.57 & 83.16 \\
PointMLP & 13.2M & 76.35 & 66.37 & 76.93 & 85.75 \\
PointMLP-Elite & 0.72M & 75.83 & 66.65 & 76.21 & 84.64 \\
RPC & 2.78M & 78.80 & 70.41 & 79.72 & \textbf{86.26} \\
\midrule
\rowcolor{gray!12}
\textbf{Mapper-GNN} & \textbf{0.60M} & 79.13 & 72.16 & 83.74 & 81.50 \\
\rowcolor{gray!12} \textbf{Mapper-GNN (8, 0.4)} & \textbf{0.60M} & \textbf{79.63} & \textbf{72.66} & \textbf{84.59} & 81.63 \\
\bottomrule
\end{tabular}
\caption{Comparison on ModelNet40-C}
\label{tab:modelnet40c_comparison}
\end{table}

\subsection{Experimental Results}
\paragraph{Robustness to Common Corruptions.}
Table~\ref{tab:modelnet40c_comparison} reports mCA and corruption-group accuracy on ModelNet40-C.
We compare Mapper-GNN with PointNet~\cite{pointnet2017}, PointNet++~\cite{pointnetpp}, DGCNN~\cite{dynamicgcnn2019}, RSCNN~\cite{liu2019rscnn},
PCT~\cite{pct}, SimpleView~\cite{goyal2021revisiting}, Point Transformer~\cite{pointtransformer}, PointNeXt~\cite{pointnext}, PointMLP~\cite{ma2022rethinking}, and RPC~\cite{pointcloudc}.

Mapper-GNN achieves the highest mCA of $79.13\%$ among the evaluated models while using only $0.6$M parameters. An alternative cover configuration,
$(n,\mathrm{overlap})=(8, 0.4)$ further increases mCA to $79.63\%$. Mapper-GNN performs particularly well under density and noise corruptions, consistent with our motivation of reducing dependence on sampling-sensitive point neighborhoods through region-level feature propagation.

The per-corruption results on ModelNet-C are reported in Table~\ref{tab:modelnet_c_per_corruption}. Mapper-GNN remains competitive across this independent benchmark and performs favorably under several point-dropping and geometric corruptions. These results suggest that its effectiveness depends partly on whether the coarse region-intersection structure remains recoverable after corruption.

\begin{table}[h]
\centering
\small
\setlength{\tabcolsep}{0.8pt}
\renewcommand{\arraystretch}{1.1}
\resizebox{\linewidth}{!}{
\begin{tabular}{lccccccc}
\toprule
\textbf{Model} & \textbf{Add-G} & \textbf{Add-L} & \textbf{Drop-G} & \textbf{Drop-L} & \textbf{Jitter} & \textbf{Rotate} & \textbf{Scale} \\
\midrule
PointNet & $13.83$ & $48.68$ & $63.66$ & $49.78$ & $60.51$ & $45.13$ & $55.65$ \\
PointNet++ & $\mathbf{70.86}$ & $62.17$ & $\mathbf{78.20}$ & $\underline{67.66}$ & $71.48$ & $64.34$ & $70.61$ \\
DGCNN & $30.90$ & $64.65$ & $74.17$ & $67.65$ & $73.77$ & \underline{$72.32$} & $71.46$ \\
RSCNN & $58.03$ & $60.32$ & $73.65$ & $50.13$ & $70.30$ & $65.67$ & $65.56$ \\
PCT & $36.85$ & $61.75$ & $73.00$ & $66.33$ & $72.77$ & $69.24$ & $67.61$ \\
SimpleView & $54.94$ & $45.80$ & $33.29$ & $36.41$ & $56.13$ & $44.38$ & $47.04$ \\
Point Transformer & $14.72$ & $37.82$ & $45.66$ & $53.81$ & $71.92$ & $68.21$ & $68.20$ \\
PointNeXt-S ($C=32$) & $46.74$ & $61.71$ & $56.48$ & $59.78$ & $65.85$ & $60.90$ & $61.33$ \\
PointNeXt-S ($C=64$) & $51.52$ & $64.22$ & $58.98$ & $61.79$ & $67.61$ & $62.55$ & $63.37$ \\
PointMLP & $36.95$ & $61.49$ & $61.47$ & $65.39$ & $71.68$ & $\mathbf{72.47}$ & \underline{$74.20$} \\
PointMLP-Elite & $27.33$ & $58.30$ & $56.64$ & $60.30$ & $71.08$ & $68.74$ & $68.79$ \\
RPC & $35.79$ & $\mathbf{68.34}$ & $70.62$ & $67.52$ & $\mathbf{76.79}$ & $71.23$ & $72.11$ \\
\midrule
\rowcolor{gray!12}
\textbf{Mapper-GNN}
& \underline{$57.02$} & \underline{$66.33$} & \underline{$74.52$} & $\mathbf{71.82}$ & \underline{$75.53$} & $61.01$ & $\mathbf{74.62}$ \\
\bottomrule
\end{tabular}
}
\caption{Per-corruption accuracy on ModelNet-C}
\label{tab:modelnet_c_per_corruption}
\end{table}

We further compare representative models on ScanObjectNN-C. As shown in Table~\ref{tab:scanobjectnn_c}, Mapper-GNN achieves the best performance under jitter, Drop-G, and Drop-L corruptions, as well as the highest clean-to-corruption retention of $71.6\%$. These results provide evidence that the proposed
representation can generalize from CAD-derived benchmarks
to real-world scanned point clouds.

\begin{table}[h]
\centering
\small
\setlength{\tabcolsep}{0.8pt}
\renewcommand{\arraystretch}{1.1}
\resizebox{\linewidth}{!}{
\begin{tabular}{lccccccc|cc}
\toprule
\textbf{Model} & \textbf{Scale} &\textbf{Jitter} &\textbf{Rotate} & \textbf{Drop-G} & \textbf{Drop-L} & \textbf{Add-G} & \textbf{Add-L} & \textbf{Retention} \\
\midrule
PointNet++ & 62.2 & 29.4 & 54.4 & 57.1 & 54.2 & 43.1 & 40.5 & 67.2 \\
DGCNN & 56.3 & 32.9 & 55.0 & 52.3 & 53.8 & 27.9 & 41.1 & 70.9 \\
PointNeXt-S ($C=64$) & \textbf{63.5} & 29.0 & 56.5 & 54.3 & 56.6 & \textbf{48.5} & \textbf{45.0} & 67.7 \\
Point Transformer & 53.1 & 30.1 & \textbf{57.7} & 31.7 & 45.7 & 24.8 & 27.3 & 60.0 \\
\midrule
\rowcolor{gray!12}
\textbf{Mapper-GNN} & 54.8 & \textbf{33.9} & 50.4 & \textbf{59.6} &  \textbf{57.7} & 38.3 & 34.0 & \textbf{71.6} \\
\bottomrule
\end{tabular}
}
\caption{Classification robustness on ScanObjectNN-C}
\label{tab:scanobjectnn_c}
\end{table}

\begin{table*}[t]
\centering
\scriptsize
\setlength{\tabcolsep}{4pt}
\renewcommand{\arraystretch}{1.0}
\begin{tabular}{lccc|ccc|c||ccc}
\toprule
\multirow{2}{*}{\textbf{Model}} & \multicolumn{3}{c|}{\textbf{5-way}} & \multicolumn{3}{c|}{\textbf{10-way}} & \multirow{2}{*} {\textbf{Retention}} & \multirow{2}{*}{\textbf{N=512}} & \multirow{2}{*}{\textbf{N=256}} & \multirow{2}{*}{\textbf{N=128}} \\
\cmidrule(lr){2-4} 
\cmidrule(lr){5-7}
& \shortstack{\textbf{1-shot}} & \shortstack{\textbf{2-shot}} & \shortstack{\textbf{3-shot}} & \shortstack{\textbf{1-shot}} & \shortstack{\textbf{2-shot}} & \shortstack{\textbf{3-shot}} & \\
\midrule
PointNet & 35.40 / 33.63 & 54.10 / 53.11 & 64.10 / 60.39 & 33.40 / 30.44 & 45.30 / 41.62 & 52.60 / 49.26 & 94.01 & \textbf{87.38} & \textbf{85.84} & \textbf{77.27} \\
PointNet++ & 46.60 / 44.15 & 58.10 / 55.67 & 64.10 / 59.41 & 37.80 / 35.14 & 47.05 / 43.24 & 54.05 / 49.25 & 93.20 & 85.92 & 61.93 & 23.14 \\
DGCNN & 56.70 / 54.93 & 70.60 / 67.62 & \underline{75.50} / 72.28 & 45.65 / 43.19 & 59.15 / 54.97 & 65.70 / 62.36 & 95.14 & 84.26 & 60.79 & 24.21 \\
RSCNN & 43.40 / 40.68 & 47.90 / 47.42 & 57.50 / 54.08 & 28.45 / 27.52 & 35.95 / 35.67 & 43.75 / 42.83 & \underline{96.77} & 80.33 & 65.47 & \underline{32.77} \\
PCT & 52.70 / 49.75 & 66.70 / 64.69 & 72.70 / 69.25 & 46.00 / 42.12 & 57.15 / 52.74 & 63.90 / 60.81 & 94.28 & 85.17 & \underline{67.29} & 22.72 \\
SimpleView & 54.50 / 49.75 & 64.10 / 58.08 & 65.60 / 58.33 & 40.30 / 34.68 & 45.50 / 39.00 & 49.95 / 42.25 & 87.86 & 55.67 & 14.48 &  6.49 \\
Point Transformer & 43.40 / 42.51 & 58.40 / 54.31 & 62.80 / 56.47 & 35.35 / 31.99 & 45.00 / 41.92 & 52.65 / 49.12 & 92.97 & 41.11 &  7.65 & -- \\
PointNeXt-S ($C=64$) & \underline{57.50} / \underline{56.85} & \underline{72.60} / \underline{69.91} & 75.00 / 70.11 & \textbf{53.15} / \textbf{49.41} & \underline{61.75} / 56.88 & \textbf{67.55} / \underline{64.24}
& 94.80 & 69.97 & 20.45 &  6.25 \\
PointMLP & 37.70 / 34.49 & 49.30 / 44.88 & 55.20 / 49.23 & 26.50 / 25.04 & 37.20 / 31.54 & 40.30 / 36.58 & 90.29 & 79.42 & 26.97 & -- \\
PointMLP-Elite & 33.40 / 32.85 & 46.80 / 44.64 & 60.60 / 53.50 & 27.90 / 24.74 & 40.15 / 35.98 & 46.75 / 40.24 & 91.06 & 79.11 & 29.07 & -- \\
RPC & 47.80 / 47.34 & 70.40 / 67.88 & \textbf{78.10} / \underline{73.53} & 44.00 / 42.69 & \textbf{63.15} / \textbf{58.99} & \underline{67.20} / 63.59
& 95.78 & 83.12 & 46.86 & -- \\
\midrule
\rowcolor{gray!12}
\textbf{Mapper-GNN} & \textbf{64.10} / \textbf{63.31} & \textbf{75.50} / \textbf{73.40} & \textbf{78.10} / \textbf{76.29} & \underline{50.50} / \underline{48.76} & 60.15 / \underline{57.55} & 66.15 / \textbf{64.35}
& \textbf{97.20} & \underline{86.47} & 66.00 & 27.86 \\
\bottomrule
\end{tabular}
\caption{Extreme few-shot classification results on clean ModelNet40 and ModelNet40-C. Each entry reports cOA\,/\,mCA (clean overall accuracy / mean corruption accuracy) and the results are averaged over 20 episodes. The last three columns indicate clean-test accuracy under reduced input point count, where `--' denotes unavailable results.}
\label{tab:fewshot_combined}
\end{table*}

\begin{figure}[h] 
\centering 
\includegraphics[width=\columnwidth]{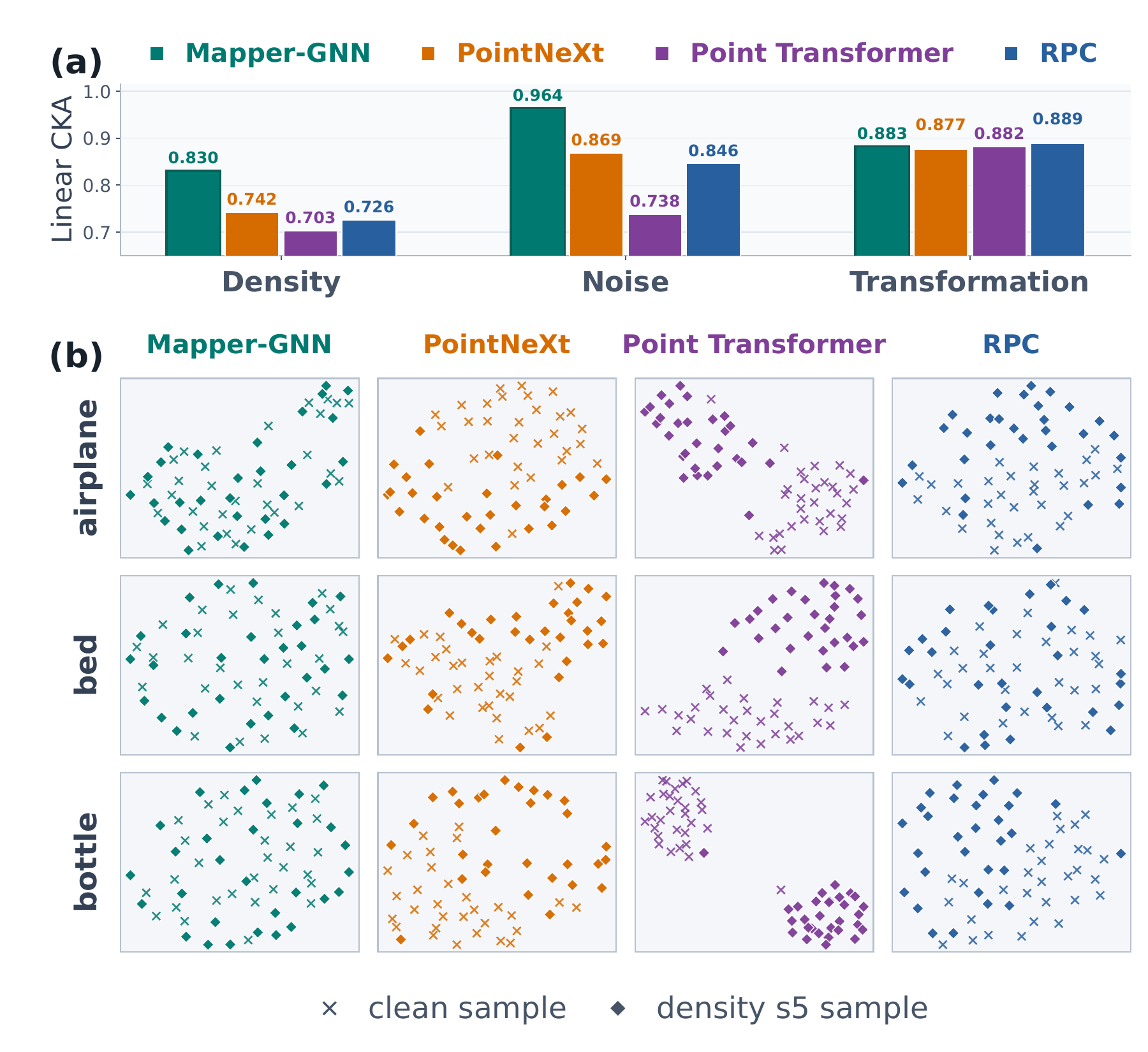 } 
\caption{Representation stability under corruption:
(a) linear CKA between clean and corrupted features.
(b) UMAP of clean and density-s5 corrupted class samples.} 
\label{fig5} 
\end{figure}

Figure~\ref{fig5} analyzes the stability of the learned representations. Mapper-GNN exhibits the highest clean-to-corrupted feature similarity under
density and noise corruptions. The corresponding UMAP visualization~\cite{mcinnes2018umap} shows substantial overlap between clean and density-s5 corrupted features. These observations provide complementary evidence that the learned representation is comparatively stable under density variation.

\begin{figure}[t]
\centering
\includegraphics[width=\columnwidth]{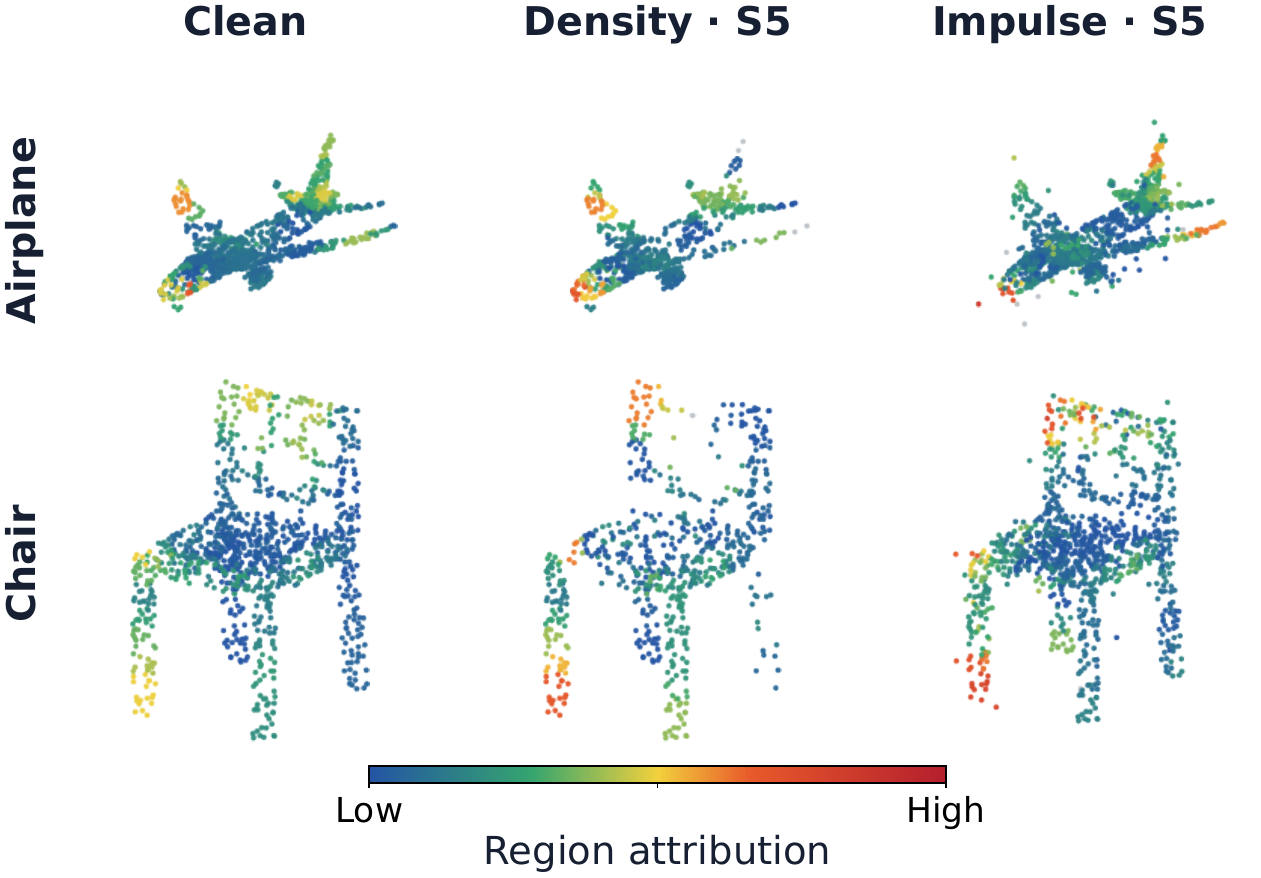}
\caption{Region attribution under clean and corruption inputs.}
\label{fig6}
\end{figure}

Figure~\ref{fig6} visualizes region-level attribution for representative airplane and chair samples under clean, density, and impulse-corrupted inputs.

\paragraph{Extreme Few-Shot Robustness.}
We conduct $n$-way $m$-shot classification experiments with $n\in\{5,10\}$ and $m\in\{1,2,3\}$, averaging results over 20 independently sampled episodes. Each selected checkpoint is evaluated on both clean ModelNet40 and ModelNet40-C. As reported in Table~\ref{tab:fewshot_combined}, Mapper-GNN achieves the highest corruption accuracy in four of the six settings and the highest average
clean-to-corruption retention of $97.20\%$. Although larger baselines obtain higher clean accuracy in some configurations, their performance degrades more under corruptions.

These results are consistent with the role of the Mapper graph as an explicit structural inductive bias. Under limited supervision, Mapper-GNN does not need to infer all geometric relations from a small number of samples; it learns over a pre-organized region graph. The strong clean-to-corruption retention therefore indicates that the learned representation remains stable beyond the sampled few-shot episodes.
Strong baselines such as PointNeXt perform well in clean few-shot classification; however, our method is more resilient to corruptions and achieves the highest mCA in four of the six few-shot settings. These findings support our hypothesis that the Mapper graph provides a useful geometric inductive bias for learning robust point-cloud representations, particularly when only a limited amount of training data is available.

\paragraph{Robustness to Global Point Sparsity.}
The last three columns in Table~\ref{tab:fewshot_combined} report clean classification accuracy as the number of input points is reduced through farthest point sampling. All results are obtained using the same checkpoints as in the main experiments.
PointNet exhibits the strongest tolerance to extreme point sparsity, while Mapper-GNN remains competitive with local-aggregation models. It achieves accuracies of $86.47\%$ at $N=512$ and $66.00\%$ at $N=256$ despite being trained with $1024$ points. These results suggest that the induced region graph can preserve useful coarse geometric structure under moderate point reduction. Performance decreases substantially at $N=128$, where severe point removal can disrupt local clusters, region intersections, and overall graph connectivity.

\subsection{Ablation Studies}

\paragraph{Model Components.}
Table~\ref{tab:component-ablation} evaluates the incremental contribution of
the main architectural components. GNN residual propagation improves mCA from
77.45\% to 77.97\%, while layer aggregation provides the largest
gain, increasing mCA to 78.87\%. Base-to-graph residual fusion further raises mCA to 79.13\%, resulting in a total improvement of 1.68 percentage points
over the baseline configuration.

\begin{table}[h] 
\centering 
\setlength{\tabcolsep}{6pt} 
\begin{tabular}{lcc} 
\toprule 
\textbf{Config} & \textbf{cOA} & \textbf{mCA} \\ 
\midrule 
None & 88.00 & 77.45 \\ 
+ GNN residual propagation & 88.36 & 77.97 \\ 
+ Layer aggregation & 88.59 & 78.87 \\ 
+ Base-to-graph residual fusion & \textbf{88.71} & \textbf{79.13} \\ 
\bottomrule 
\end{tabular} 
\caption{Incremental ablation study} 
\label{tab:component-ablation} 
\end{table}

% \paragraph{Graph Depth.}
% Table~\ref{tab:gcn-layer-ablation} examines the number of message-passing layers in the default graph encoder. Increasing the depth from one to two GCN layers substantially improves both clean overall accuracy (cOA) and mCA. Additional layers provide only marginal gains, with the two-layer model performing comparably to the four-layer model. This suggests that most relevant structural interactions in the Mapper graph can be captured within a small number of message-passing steps.

% \begin{table}[h] 
% \centering 
% \setlength{\tabcolsep}{12pt} 
% \begin{tabular}{ccc} 
% \toprule 
% \textbf{\# GCN layers} & \textbf{cOA} & \textbf{mCA} \\ 
% \midrule 
% 1 & 85.53 & 74.41 \\ 
% 2 & 88.34 & 78.82 \\ 
% 3 & 88.37 & 78.81 \\ 
% 4 & \textbf{88.71} & \textbf{79.13} \\ 
% \bottomrule 
% \end{tabular} 
% \caption{Effect of the number of GCN layers} 
% \label{tab:gcn-layer-ablation} 
% \end{table}

\paragraph{Graph Encoder and Lens.}
Table~\ref{tab:mapper_ablation} compares GIN, GCN, and GAT encoders using the identity and PCA$_3$ lenses. GAT with PCA$_3$ achieves the highest mCA of 79.6\%, whereas GCN offers a more favorable robustness--complexity trade-off with fewer than half as many parameters. PCA$_3$ generally matches or slightly outperforms the identity lens, indicating that Mapper-GNN is not restricted to a particular graph encoder or lens function.

\begin{table}[h] 
\centering 
\setlength{\tabcolsep}{1.4pt} 
\renewcommand{\arraystretch}{1.1} 
\begin{tabular}{ll|ccccc} 
\toprule 
\textbf{Encoder} & \textbf{Lens} & Params & Density & Noise & Trans. & \textbf{mCA} \\ 
\midrule 
GIN & Identity & 0.87M & $\mathbf{74.1}$ & 78.2 & 80.5 & 77.6 \\ 
& PCA$_3$ & 0.87M & 71.2 & 82.3 & 81.2 & 78.2 \\ 
\midrule 
GCN & Identity$^\dagger$ & $\mathbf{0.60M}$ & 72.2 & 83.7 & 81.5 & 79.1 \\ 
& PCA$_3$ & $\mathbf{0.60M}$ & 71.6 & 83.7 & $\mathbf{82.3}$ & 79.2 \\ 
\midrule 
GAT & Identity & 1.40M & 73.0 & $\mathbf{84.1}$ & 81.3 & 79.4 \\ 
& PCA$_3$ & 1.40M & 72.9 & 83.8 & 82.2 & $\mathbf{79.6}$ \\ 
\bottomrule 
\end{tabular} 
\caption{Ablation study of Mapper-based variants on ModelNet40-C. $^\dagger$ denotes the representative configuration used in the main comparison.} 
\label{tab:mapper_ablation} 
\end{table}

\paragraph{Mapper Cover Parameters.}
Table~\ref{tab:mapper_cover_ablation} evaluates the effects of cover resolution and overlap ratio. The coarse configuration with $n=4$ consistently underperforms, whereas configurations with $n=6$ and $n=8$ achieve stable mCA values ranging from 78.58\% to 79.63\%. The best performance is obtained with $(n,\mathrm{overlap})=(8,0.4)$. Overall, the results remain stable across several cover configurations, suggesting that the observed robustness does not depend on a narrowly selected set of Mapper parameters.

\begin{table}[h]
\centering
{\small
\setlength{\tabcolsep}{3.8pt}
\renewcommand{\arraystretch}{1.08}
\begin{tabular*}{\columnwidth}{ @{\extracolsep{\fill}} c cc cc cc @{}}
\toprule
\textbf{Cover $n$} & \multicolumn{2}{c}{\textbf{Overlap $=0.2$}} & \multicolumn{2}{c}{\textbf{Overlap $=0.3$}} & \multicolumn{2}{c} {\textbf{Overlap $=0.4$}} \\
\cmidrule(lr){2-3}
\cmidrule(lr){4-5}
\cmidrule(lr){6-7}
& \textbf{cOA} & \textbf{mCA} & \textbf{cOA} & \textbf{mCA} & \textbf{cOA} & \textbf{mCA} \\
\midrule
4 & 88.70 & 77.98 & 88.83 & 77.45 & 88.91 & 76.85 \\
6 & 88.63 & 79.30 & 88.74 & 79.13$^\dagger$ & \textbf{89.10} & 78.74 \\
8 & 87.74 & 78.58 & 88.88 & 79.40 & 88.95 & \textbf{79.63} \\
\bottomrule
\end{tabular*}
}
\caption{Effect of Mapper cover resolution and overlap ratio.
$^\dagger$ denotes the default setting $(n,\mathrm{overlap})=(6,0.3)$.}
\label{tab:mapper_cover_ablation}
\end{table}

\section{Conclusion}

We investigated whether a Mapper-induced region graph can provide an effective structural inductive bias for robust point cloud classification. Mapper-GNN organizes each input point cloud into overlapping, locally coherent regions and performs feature propagation over their intersection graph. By combining lightweight point-to-node encoding, residual multi-level message passing, and base-to-graph feature fusion, the proposed framework achieves competitive corruption robustness with only $0.6$M parameters, without corruption-specific augmentation, model ensembling, or test-time adaptation. Its strongest results under density and noise corruptions, together with favorable performance in few-shot and reduced-point settings, demonstrate the potential of region-level structural abstraction for robust and parameter-efficient point cloud learning.

The effectiveness of the representation nevertheless depends on the recoverability of the induced region structure. Severe point removal can disrupt local clusters, region intersections, and graph connectivity, leading to degraded performance. Moreover, although the neural classifier is lightweight, Mapper graph construction introduces additional preprocessing cost and remains the primary computational bottleneck. Future work will focus on accelerating graph construction and developing adaptive or learnable Mapper components, including learnable lenses, differentiable covers, and soft region assignments. These extensions may improve computational efficiency and enable the region graph to adapt more effectively to severe sampling and structural variations.

\bibliography{aaai2027}


\end{document}